\documentclass[]{iscram}

\usepackage[utf8]{inputenc}

\usepackage{algorithm}
\usepackage{algorithmic}
\usepackage{amsmath}
\usepackage{bbm}

\usepackage{booktabs}
\usepackage[normalem]{ulem}
\useunder{\uline}{\ul}{}

\usepackage{lipsum}
\usepackage{tikz}
\usepackage{fancyvrb}
\usepackage{listings}
\usepackage{enumitem}
\usepackage{tcolorbox}
\usepackage{multicol}
\usepackage{siunitx}
\usepackage{xpatch}
\lstdefinestyle{common}{
  xleftmargin=.5em,
  xrightmargin=.5em,
  frame=single,framesep=.5em,framerule=0pt,
  fancyvrb=true,
  basicstyle=\ttfamily,
  keywordstyle=\color{cyan!50!blue!75!black}\bfseries,
  commentstyle=\color{red!50!black}\itshape,
  stringstyle=\ttfamily\color{green!50!black},
  numbers=none,
  showspaces=false,
  showstringspaces=false,
  fontadjust=true,
  keepspaces=true,
  flexiblecolumns=true,
  emphstyle=\color{red},
}
\lstdefinestyle{TeX}{
  style=common,
  backgroundcolor=\color{blue!5},
  aboveskip=5pt,
  belowskip=5pt,
  language=[LaTeX]TeX,
  moretexcs={
    abstract, addbibresource, iscramset, keywords, mainmatter,
    maketitle, printbibliography, subsection, subsubsection, url,
    urldef, href, includegraphics, ldots, parencite, citeauthor,
    citeyear, citetitle, midrule, toprule, bottomrule
  },
  fancyvrb=true,
}
\lstdefinestyle{console}{
  style=common,
  backgroundcolor=\color{gray!10},
  aboveskip=5pt,
  belowskip=5pt,
}

\newlist{options}{description}{1}
\setlist[options]{%
  beginpenalty=10000,%
  itemsep=.5\parskip plus .3\parskip minus .2\parskip,
  parsep=.5\parskip plus .3\parskip minus .2\parskip,
  topsep=.5\parskip plus .3\parskip minus .2\parskip,
  partopsep=.5\parskip plus .3\parskip minus .2\parskip,
  style=nextline,labelindent=1em,%
  font=\normalfont\ttfamily}

\colorlet{macro color}{cyan!50!blue!75!black}
\colorlet{option color}{red!50!black}
\colorlet{generic color}{green!40!black}

\newtcolorbox{pseudoTeX}{colback=blue!5,colframe=blue!5,before=\nobreak}
\let\LaTeXorig\LaTeX
\renewcommand\LaTeX{\bgroup\fontfamily{lmr}\selectfont\upshape\LaTeXorig\egroup}

\urldef{\sitewebgind}\url{http://gind.mines-albi.fr}
\urldef{\sitewebminesalbi}\url{http://www.mines-albi.fr}
\urldef{\gmailpaulgaborit}\url{email adress}
\urldef{\jdmail}\url{...@example.com}
\urldef{\sitex}\url{www.example.com}

\urldef{\hpzmail}\url{pzou3@uic.edu}
\urldef{\ccmail}\url{cornelia@uic.edu}
\urldef{\yzmail}\url{yzhou232@uic.edu}
\urldef{\dcmail}\url{dcaragea@ksu.edu}

\iscramset{
  title={
    Semi-Supervised Few-Shot Learning for Fine-Grained Disaster Tweet Classification\\
  },
  short title={CrisisMatch},
  author={
    short name={Peng},
    full name=Henry Peng Zou,
    affiliation={
      University of Illinois Chicago\\
       \href{mailto:pzou3@uic.edu}{\hpzmail}
    },
  },
  author={
    full name={Cornelia Caragea},
    affiliation={
      University of Illinois Chicago\\
       \href{mailto:cornelia@uic.edu}{\ccmail}
    },
  },
  author={
    full name={Yue Zhou},
    affiliation={
      University of Illinois Chicago\\
       \href{mailto:yzhou232@uic.edu}{\yzmail}
    },
  },
  author={
    full name={Doina Caragea},
    affiliation={
      Kansas State University\\
       \href{mailto:dcaragea@ksu.edu}{\dcmail}
    },
  },
  WiP Paper 2023=Social Media for Crisis Management
}

\usepackage{tikzpagenodes}
\usetikzlibrary{fit}

\begin{document}

\maketitle

% \makeatletter
% {\centering\large\iscram@version{}\\\iscram@date\par}
% \makeatother

%%%%%%% Abastract  %%%%%%%%%%%

\abstract{

The shared real-time information about natural disasters on social media platforms like Twitter and Facebook plays a critical role in informing volunteers, emergency managers, and response organizations. However, supervised learning models for monitoring disaster events require large amounts of annotated data, making them unrealistic for real-time use in disaster events. %Current semi-supervised approaches also over-assume the number of available labels and provide limited information through binary classifications. 
To address this challenge, we present a fine-grained disaster tweet classification model under the semi-supervised, few-shot learning setting where only a small number of annotated data is required. Our model, CrisisMatch, effectively classifies tweets into fine-grained classes of interest using few labeled data and large amounts of unlabeled data, mimicking the early stage of a disaster. Through integrating effective semi-supervised learning ideas and incorporating TextMixUp, CrisisMatch achieves %over +9.5\% and +12.9\% 
performance improvement on two disaster datasets of 11.2\% on average. Further analyses are also provided for the influence of the number of labeled data and out-of-domain results.
}


% \newpage

%%%%%%%%%%%% Introduction %%%%%%%%%%%%%

\section{1.\quad Introduction}
% background and motivation, challenges, limitations of current state-of-the-art approaches
% Semi-supervised Few-Shot Learning for Fine-grained Crisis Tweet Classification

In times of natural disasters, individuals share content, facts, recommendations, and warnings about the disaster in real-time on social media platforms such as Twitter and Facebook. Such information is crucial to help volunteers, emergency managers, and response organizations become more situationally aware and efficient in their rescue activities \parencite{varga2013aid, vieweg2014integrating}. 

Although existing works leverage such information to build models to monitor disaster events, many approaches require annotating large amounts of data when disasters happen, which are unrealistic due to the limited response time \parencite{caragea2016identifying, chowdhury2020cross}. Current semi-supervised approaches also \emph{over-assume} the number of available labels (e.g., more than 50 labels per class) \parencite{alam2018graph, karisani2021semi, sirbu2022multimodal}, which is hard to obtain when the number of classes is large. In addition, while the popular coarse, binary classification identifying whether a tweet is disaster-relevant can be useful, it is more informative to have fine-grained, multi-categorical classifications providing disaster information from different angles \parencite{plotnick2015red, reuter2018social, imran2020using, alam2021humaid}.

To address the above challenges, we investigate the problem of fine-grained disaster tweet classification under the semi-supervised, few-shot learning setting. It aims to classify tweets during a disaster event into fine-grained classes of interest (e.g., injured or dead people, infrastructure and utility damage, caution and advice, rescue volunteering, or donation effort) \parencite{alam2021humaid}. In addition, in our few-shot setting, we only utilize five labeled data points per class, with the rest data being unlabeled and naturally imbalanced. This setting mimics the early stage of a disaster, which is also the most valuable time for rescue and escape.

This paper first studies the effectiveness of various semi-supervised learning components on leveraging few labeled data and large amounts of unlabeled data for disaster tweet classification. Then, we extend the pseudo-labeling algorithm \parencite{lee2013pseudo, zhang2021flexmatch} through entropy minimization, data augmentation, and consistency regularization. Particularly, we incorporate TextMixUp \parencite{zhang2018mixup, chen-etal-2020-mixtext}, which encourages models to behave linearly among samples and avoid overfitting. Then, we propose CrisisMatch, by combining those effective components for fine-grained disaster tweet classification in the semi-supervised few-shot setting. Experimental results show that the proposed CrisisMatch achieves over 11.2\% performance improvement on average on two disaster datasets. We also provide further analyses for CrisisMatch on the influence of the number of labeled data and out-of-domain results.

% Our contributions are summarized as follows:

% Summary of contributions

\section{2.\quad Related Work}
% a discussion of related work

\textbf{Semi-supervised learning. }
Semi-supervised learning is developed to alleviate the reliance on labeled data by leveraging unlabeled data to train more robust models \parencite{lee2013pseudo, berthelot2019mixmatch, xie2020self, zhang2021flexmatch}. Self-training adopts the idea of using the output probability of the model as a soft label for unlabeled data \parencite{scudder1965probability, mclachlan1975iterative, lee2013pseudo, xie2020self}. Pseudo-labeling modifies self-training by using hard labels, instead of soft labels, and confidence thresholding to reduce confirmation bias and select high-quality pseudo-labels for training \parencite{lee2013pseudo, zhang2021flexmatch}. Mean Teacher \parencite{tarvainen2017mean} proposes to use the exponential moving average of model weights for label predictions on unlabeled data. MixMatch \parencite{berthelot2019mixmatch} utilizes sharpening to encourage low-entropy prediction for unlabeled data and uses MixUp \parencite{zhang2018mixup} to mix labeled and unlabeled data. MixText \parencite{chen-etal-2020-mixtext} adapts MixUp to text settings by interpolating hidden representations of texts.

\textbf{Disaster tweet classification. }
Disaster tweet classification has made huge progress in recent years to improve crisis relief operations. \parencite{imran2013practical, imran2015processing} proposed to classify disaster tweets to obtain useful information for disaster understanding and rescue. \parencite{nguyen2017robust} introduced Convolutional Neural Networks (CNNs) to classify informative disaster-related tweets. \parencite{kruspe2019detecting} studied the supervised few-shot learning setting, where only a few labeled data are used for training a disaster tweet classifier. \parencite{li2021combining} combined self-training with CNN and BERT pre-trained language models to improve the performance of classifying disaster tweets where only unlabeled data is available. Other works \parencite{alam-etal-2018-domain, mazloom2019hybrid, li2018disaster} explored unsupervised domain adaptation in which only unlabeled data is available for the current crisis event while labeled data is available from previous disaster events. However, most prior works either focus on coarse, binary classification or assume that many labeled data is available. In contrast, this work studies fine-grained disaster tweet classification under the few-shot setting, which we believe can benefit the early stage of disaster analysis and rescue.

\section{3.\quad Method}

This section first reviews several classical semi-supervised learning ideas and components, including self-training, entropy minimization, and consistency regularization, then describes TextMixUp and our CrisisMatch algorithm for the task of fine-grained disaster tweet classification in the few-shot setting.

% 1. PSL, Entropy Minimization, Sharpening \\
% 2. Aug, Consistency Regularization \\
% 3. MixUp \\ 
% (4. CrisisMatch)

\subsection{3.1 Self-Training and Entropy Minimization}

In the semi-supervised few-shot learning setting, there are usually only a few labeled data but a large amount of unlabeled data. Self-training takes advantage of unlabeled data by using the model itself to infer predictions on unlabeled data, and then utilizing these predictions as pseudo-labels for training \parencite{scudder1965probability, mclachlan1975iterative, lee2013pseudo, xie2020self}. However, this may result in the issue of confirmation bias: If these pseudo-labels are incorrect and the model is trained on them, the model can become worse and worse, continually confirming its own incorrect bias. A common strategy to alleviate this problem is to use a threshold to select only pseudo-labels whose largest class probability surpasses the threshold. 

Besides, a basic assumption in many semi-supervised learning methods is that data in the same class are clustered together and thus a good classifier's decision boundary between classes should not pass through high-density regions of the data manifold. Entropy minimization achieves this by encouraging the model to produce low-entropy predictions on unlabeled data \parencite{grandvalet2004semi, miyato2018virtual}. This is because the model will by necessity output high-entropy predictions for some samples if the decision boundary falls in high-density regions. Here we introduce two ways to implement entropy minimization: Pseudo-labeling and Sharpening. 

Pseudo-Labeling \parencite{lee2013pseudo, zhang2021flexmatch} uses hard (i.e., one-hot) labels from high-confidence predictions on unlabeled data as the targets for training, which implicitly enforces the model to output low-entropy predictions. Formally, pseudo-labeling minimizes the following loss function: 
% [[Formula: Pseudo Labeling]]
\begin{equation}
    \frac{1}{\mu B}\sum_{b=1}^{\mu B} \mathbbm{1}(\max(p_m(y|\hat{u}_{b}))>\tau)L(q_m(y|\hat{u}_{b}),p_m(y|\hat{u}_{b})) 
\label{eqn: pseudo_labeling}
\end{equation}
where $\mu$ is the ratio of unlabeled data to labeled data, $B$ is the batch size of labeled data, $\hat{u}_{b}$ is a \textbf{\textit{stochastically}} data augmentation function, $u_b$ represents an unlabeled sample, $p_m$ denotes model's probability prediction, $L$ is L2 loss or cross-entropy loss, $\tau$ is a fixed threshold and $q_m(y|\hat{u}_{b})$ is the hard one-hot pseudo-label.

Sharpening in MixMatch \parencite{berthelot2019mixmatch} uses soft labels and explicitly decreases the entropy of predicted label distribution on unlabeled data by adjusting all probabilities to the power of $1/T$ and normalizing: 
% [[Formula: Sharpening]] 
\begin{equation}
    \text{Sharpen}(p_i, T) = \frac{p_i^{\frac{1}{T}}}{ \sum_{j = 1}^C p_j^{\frac{1}{T}}}
\label{eqn:sharpen}
\end{equation}
where $T$ is the temperature hyperparameter and $C$ is the number of total classes. Lowering the temperature hyper-parameter $T$ encourages the model to produce low-entropy predictions. When $T \rightarrow 0$, the predicted label becomes a one-hot hard label.

\subsection{3.2 Data Augmentation and Consistency Regularization}

Data augmentation is a common technique to alleviate overfitting and help regularize models, especially when using smaller datasets. It artificially increases the amount of training data by generating perturbed inputs with transformations that are assumed not to change original class semantics. For instance, synonym replacement, random swap, random insertion and random deletion are convenient and easy data augmentations for text \parencite{wei2019eda}. More advanced techniques such as back-translation \parencite{fadaee-etal-2017-data, sugiyama2019data} and paraphrasing \parencite{kumar2019submodular} are also proposed to generate more diversified augmented data but are more costly and complicated to implement. 

Consistency regularization leverages the idea that a classifier should have similar predictions for data before and after augmentation \parencite{bachman2014learning, sajjadi2016regularization}. A straightforward implementation is to add the loss term: 
% [Formula: Consistency regularization] 
\begin{equation}
    \sum_{b=1}^{\mu B} ||p_m(y|\hat{u}_{b})-p_m(y|\hat{u}_{b})||_2^2 
\label{eqn: consistency_regularization}
\end{equation}
Another implicit way is to use the average prediction of several different augmentations of an unlabeled sample as the common pseudo-label of all augmented data \parencite{berthelot2019mixmatch}. Formally, 
% [Formula: average predictions]
\begin{equation}
    q_m(y|\hat{u}_{b}) = \frac{1}{K}\sum_{k}^{K} p_m(y|\hat{u}_{b,k}))
\label{eqn: avg_pred}
\end{equation}
where $K$ is the number of augmentations, $\hat{u}_{b,k}$ is the k-th augmented data, and $q_m(y|\hat{u}_{b})$ is the common pseudo-label for all the corresponding augmented unlabeled data.

\subsection{3.3 MixUp and TextMixUp}
MixUp is a technique proposed by \parencite{zhang2018mixup} that can regularize the model to behave linearly among training samples and alleviate overfitting. The idea is to generate numerous new virtual training samples by linearly mixing two randomly sampled input and their one-hot labels: 
% [Formula: Mixup]. 
\begin{align}
\tilde{x}  = \lambda {x}_i + (1-\lambda){x}_j  
\label{eqn: MixUp_x} \\
\tilde{{y}} = \lambda {y}_i + (1-\lambda){y}_j 
\label{eqn: MixUp_y}
\end{align}
However, applying MixUp directly on text input seems infeasible since the interpolation of discrete text tokens makes no sense. 

A more practical method is to interpolate hidden representations of texts at a certain layer and use the mixed representation for future layer and model prediction \parencite{verma2019manifold, chen-etal-2020-mixtext}:
% [[Formula: TextMixup]]. 
% \begin{align}
% {h}_l^i =& g_l({h}_{l-1}^i), l\in[1, m] \\
% {h}_l^j =& g_l({h}_{l-1}^j), l\in[1, m]
% \end{align}
\begin{align}
\tilde{{h}}_m =  \lambda {h}_m^i + (1-\lambda) {h}_m^j \\
\tilde{{h}}_l =  g_l(\tilde{{h}}_{l-1}),  l \in[m+1, L]
\end{align}
where ${h}_m^i$ are the hidden representation of $m$-th layer for sentence $i$, $g_l(\cdot)$ is the encoder function. 

We refer to the above method as TextMixUp and it has the potential to create many more virtual training samples since it can interpolate representation at any layer of the encoder instead of just the input samples in the original MixUp.

\subsection{3.4 Our Algorithm: CrisisMatch}

%%%%%%%%%%%%%%%%%% Algorithm %%%%%%%%%%%%%%%%%%
\begin{algorithm}[t]
\caption{CrisisMatch algorithm.}
\label{alg:CrisisMatch}
\begin{algorithmic}[1]

\STATE \textbf{Input:} Labeled batch $\mathcal{X}=\{(x_b,y_b):b \in (1, 2, \dots, B)\}$, unlabeled batch $\mathcal{U}=\{u_b:b \in (1, 2, \dots, \mu B)\}$, unsupervised loss weight $w_u$, confidence threshold $\tau$, number of augmentations $K$.
% Hard Pseudo-Labeling
\STATE $\hat{\mathcal{U}} = \{ \, \}$
\FOR{$b = 1$ \TO $B$}
\STATE $\hat{x}_b = \text{Aug}(x_b)$ \label{line:augment_labeled}
\color{gray}\COMMENT{\textit{Data augmentation for labeled examples}}\color{black} 
   \FOR{$k = 1$ \TO $K$}
   \STATE $\hat{u}_{b, k} = \text{Aug}(u_b)$   \label{line:augment_unlabeled}
   \color{gray}\COMMENT{\textit{Data augmentation for unlabeled examples}}
   \ENDFOR
   \STATE $q_b = \frac{1}{K}\sum_k p_m(y \mid \hat{u}_{b, k})$ \label{line:avg_pred}
   \color{gray}\COMMENT{\textit{Compute average prediction across different augmentations of $u_b$} as guessed probability distribution target for all $\hat{u}_{b, k}$}\color{black}
   \color{blue}
   \IF{$\max(q_b) \geq \tau$}
   \STATE $\hat{q}_b = \arg\max(q_b)$  \label{line:pseudo_label}
   \color{gray}\COMMENT{\textit{Compute one-hot guessed labels from high-confidence predictions}}\color{blue}
   \STATE $\hat{\mathcal{U}} \leftarrow (\hat{u}_{b, k}, \hat{q}_b) $
   \color{gray}\COMMENT{\textit{Use guessed labels as training targets for augmented unlabeled examples}}\color{blue}
   \ENDIF
\ENDFOR
% TextMixUp
\STATE $\hat{\mathcal{X}} = \big((\hat{x}_b, y_b); b \in (1, \ldots, B)\big)$
\color{gray}\COMMENT{\textit{Augmented labeled examples and their labels}}\color{black}
\STATE $\mathcal{W} = \text{Shuffle}\big(\text{Concat}(\hat{\mathcal{X}}, \hat{\mathcal{U}})) \big) $
\color{gray}\COMMENT{\textit{Shuffle all labeled and unlabeled data}}\color{black}
\color{blue}
\STATE $\tilde{\mathcal{X}} = \big(\text{TextMixUp}(\hat{\mathcal{X}}_i, \mathcal{W}_i); i \in (1, \ldots, |\hat{\mathcal{X}}|)\big) $ \label{line:TextMixUp}
\color{gray}\COMMENT{\textit{Apply TextMixUp to labeled data and entries from $\mathcal{W}$}}\color{blue}
\STATE $\tilde{\mathcal{U}} = \big(\text{TextMixUp}(\hat{\mathcal{U}}_i, \mathcal{W}_{i+|\hat{\mathcal{X}}|}); i \in (1, \ldots, |\hat{\mathcal{U}}|)\big) $
\color{gray}\COMMENT{\textit{Apply TextMixUp to unlabeled data and the rest of $\mathcal{W}$}}\color{blue}
% Compute Loss
\color{black}
\STATE $\mathcal{L}_x = \frac{1}{|\tilde{\mathcal{X}|}} \sum_{\tilde{x}, \tilde{y} \in \tilde{\mathcal{X}}} H(\tilde{y}, p_m(y \mid \tilde{x}))$
\color{gray}\COMMENT{\textit{Compute loss for labeled data}}\color{black} \label{line:loss_labeled}
\STATE $\mathcal{L}_u = \frac{1}{|\tilde{\mathcal{U}}|} \sum_{\tilde{u}, \tilde{q} \in \tilde{\mathcal{U}}} \|\tilde{q} - p_m(y \mid \tilde{u})\|_2^2$ \label{line:loss_unlabeled}
\color{gray}\COMMENT{\textit{Compute loss for unlabeled data}}\color{black}
\STATE \textbf{Return:} $\mathcal{L}_{x} + w_u\mathcal{L}_{u}$
\end{algorithmic}
\end{algorithm}
%%%%%%%%%%%%%%%%%% Algorithm %%%%%%%%%%%%%%%%%%

In this section, we introduce our algorithm CrisisMatch, which incorporates the components and ideas described above for the task of semi-supervised few-shot disaster tweet classification. The complete algorithm for CrisisMatch is presented in Algorithm \ref{alg:CrisisMatch}.

Given a labeled batch $\mathcal{X}=\{(x_b,y_b):b \in (1, 2, \dots, B)\}$ and a unlabeled batch $\mathcal{U}=\{u_b:b \in (1, 2, \dots, \mu B)\}$. We first apply \textbf{\textit{data augmentation}} to both labeled and unlabeled data. Specifically, we generate one augmented labeled sample $\hat{x}_b $ and $K$ augmented unlabeled samples $\hat{u}_{b, k}$ (algorithm \ref{alg:CrisisMatch}, line \ref{line:augment_labeled}, \ref{line:augment_unlabeled}). Then we implicitly enforce \textbf{\textit{consistency regularization}} by using the average prediction across different augmentation of $u_b$ as the guessed probability distribution target for all $K$ augmented unlabeled samples $\hat{u}_{b, k}$ (algorithm \ref{alg:CrisisMatch}, line \ref{line:avg_pred}). 

To encourage \textbf{\textit{entropy minimization}}, we use hard pseudo-labeling by computing one-hot pseudo-labels from unlabeled augmented data that receive high-confidence predictions (algorithm \ref{alg:CrisisMatch}, line \ref{line:pseudo_label}). \textbf{\textit{TextMixUp}} is then applied on shuffled labeled and unlabeled data to further regularize the model to behave linearly between samples and effectively leverage limited labeled data (algorithm \ref{alg:CrisisMatch}, line \ref{line:TextMixUp}). Lastly, we compute cross-entropy loss for labeled data and compute $L_2$ loss for unlabeled data since $L_2$ loss is bounded for probabilities and less sensitive to wrong predictions of unlabeled data (algorithm \ref{alg:CrisisMatch}, line \ref{line:loss_labeled}, \ref{line:loss_unlabeled}).

Our algorithm CrisisMatch differs from MixMatch\parencite{berthelot2019mixmatch} mainly as follows: CrisissMatch uses hard pseudo-labeling for entropy minimization instead of sharpening in MixMatch. We empirically found that hard-pseudo-labeling achieves better results than sharpening on both disaster datasets, as shown in Section 4.5 and Table \ref{tab6:entropy_min}. Besides, MixMatch uses all sharpened guessed labels on unlabeled for training, while CrisisMatch only selects high-confidence predictions as pseudo-labels for training. We argue that not all guessed labels on unlabeled data should be used since many of them may be incorrect, especially in the few-shot setting, and degenerate model performance. Lastly, CrisisMatch leverages TextMixUp rather than MixUp since TextMixUp can interpolate text hidden representations in any layer of encoder while MixUp only interpolates in input space and is not feasible for interpolating discrete text tokens.

% 1. Augmentation
% 2. Average predictions
% 3. Label-Guessing/Pseudo-labeling
% 4. TextMixUp
% 5. Compute Loss for labeled and unlabeled data

\section{4 Experiments}

\subsection{4.1 Datasets}

To evaluate the performance of our proposed methods and baseline methods for fine-grained disaster tweet classification, we use three datasets sampled and processed from HumAID(\parencite{alam2021humaid}: 1) Earthquake; 2) Wildfires; 3) Floods. We use Earthquake and Wildfires datasets for in-domain evaluation and Floods dataset for out-of-domain evaluation.

These datasets comprise tweets collected in natural disasters that occurred between 2016 and 2019, such as the 2018 California Wildfires and the 2019 Pakistan Earthquake. Originally, there are 10 classes for each disaster type. However, some classes have less than 100 data and even less than 10 data after splitting, and are difficult for effective evaluation. Therefore, we discard those classes with less than 100 data and use the 7-class version of the datasets for a more convincing evaluation. Table \ref{tab1:label_distribution} shows all labels and class-wise distributions of the datasets used in our experiments. Table \ref{tab2:data_splits} shows the data splits for each used disaster dataset. We will release our sampled and processed datasets to make them convenient to use for future researchers.

%%%%%%%%%%% Table  %%%%%%%%%%%
% \usepackage{booktabs}
\begin{table}[H]
\centering
\begin{tabular}{lrrrr}
\toprule
\textbf{Labels} & \textbf{Wildfires} & \textbf{Earthquake} & \textbf{Floods} \\ 
\midrule
caution\_and\_advice  & 245 & 629 & 246 \\
infrastructure\_and\_utility\_damage  & 673 & 728 & 453\\
injured\_or\_dead\_people   & 1946 & 1489 & 409\\
not\_humanitarian   & 1397 & 498 & 865\\
other\_relevant\_information  & 1349 & 707 & 1284\\
rescue\_volunteering\_or\_donation\_effort   & 2349 & 2049 & 5401 \\
sympathy\_and\_support  & 633 & 2520 & 1040\\
\midrule
total   & 8592 & 8620 & 9698\\ 
\bottomrule
\end{tabular}
\caption{Labels distribution for each dataset.}
\label{tab1:label_distribution}
\end{table}

\begin{table}[H]
\centering
\begin{tabular}{lrrrr}
\toprule
\textbf{Datasets} & \textbf{Size} & \textbf{Train(80\%)} & \textbf{Dev(10\%)} & \textbf{Test(10\%)} \\ 
\midrule
Wildfires & 8592 & 6874 & 859 & 859 \\
Earthquake & 8620 & 6896 & 862 & 862 \\
Floods & 9698 & 7758 & 970 & 970 \\
\bottomrule
\end{tabular}
\caption{Data splits for each disaster dataset.}
\label{tab2:data_splits}
\end{table}

%%%%%%%%%%% Table  %%%%%%%%%%%

\subsection{4.2 Evaluation Setting}
We use accuracy and macro-F1 as our evaluation metrics. Accuracy is used to measure the overall performance of different approaches on all testing data and macro-F1 is used because it takes class imbalance setting into account and measures the average performance of all classes. All methods are evaluated on the 5-shot setting in default: only 5 labeled data are randomly sampled from the training set and used, and the rest of the training data are treated as unlabeled data. We report accuracy and macro-F1 averaged across three runs with the same three random seeds for all methods.

\subsection{4.3 Experimental Setup}
To test the effectiveness of the proposed semi-supervised learning methods and components discussed in Section 3, we perform experiments for the following methods:
1) Supervised Baseline: supervised baseline that uses the pre-trained BERT-base-uncased model with one classification layer added and \textit{fine-tuned only on labeled data} for our classification task. For a fair comparison, all methods used the same model architecture with this supervised baseline.
2) PSL: plain pseudo-labeling that uses not only labeled data but also unlabeled data with high-confidence predictions and their hard pseudo-labels for training; 
3) PSL++: add data augmentations and consistency regularization to PSL; 
4) TextMixUp: apply TextMixUp introduced in Section 3 to the supervised baseline; 
5) CrisisMatch: our proposed algorithm, as described in Section 3. 
Note that Supervised Baseline and TextMixUp use only labeled data (e.g., 5 labeled data per class in default) and other approaches utilize both labeled data and unlabeled data.

% Implementation Details
All methods use the same model architecture BERT-base-uncased model and hyper-parameters in default. We set batch size as 32, learning rate as 2e-5, and use AdamW as optimizer. The maximum sequence length is 64. $\lambda$ in TextMixUp are sampled from the Beta distribution with sampling hyper-parameter $\alpha=0.75$. The weight of the unlabeled loss is searched among \{0.1, 0.5, 1, 5, 10, 50, 100\} and set to 10 in default. We adopt linear ramps-up strategy for the unlabeled loss weight and set ramps-up length as 1000 iterations. Confidence threshold $\tau$ is set to 0.75 and sharpening temperature $T$ is set to 0.5. We use synonym replacement and random swap as augmentation methods and set $K$, the number of augmentations for unlabeled samples, to 2.

\subsection{4.4 Main Results}

% \usepackage{booktabs}
\begin{table}[H]
\centering
\begin{tabular}{@{}lll|ll@{}}
\toprule
\textbf{Datasets} & \multicolumn{2}{c|}{\textbf{Wildfires}} & \multicolumn{2}{c}{\textbf{Earthquake}} \\ \midrule
\textbf{Methods} & \textbf{Accuracy} & \textbf{Macro-F1} & \textbf{Accuracy} & \textbf{Macro-F1} \\ \midrule
Supervised Baseline & 53.9 ± 1.3 & 44.2 ± 1.8 & 51.1 ± 2.0 & 41.4 ± 1.9 \\ \midrule
PSL & 59.3 ± 0.5 & 42.2 ± 2.4 & 57.5 ± 4.9 & 39.0 ± 3.4 \\
PSL++ & 60.1 ± 5.3 & 46.8 ± 2.5 & 58.2 ± 1.7 & 41.8 ± 3.6 \\
TextMixUp & 56.0 ± 1.7 & 46.2 ± 1.6 & 58.1 ± 5.5 & 46.5 ± 5.9 \\
CrisisMatch & \textbf{63.4 ± 1.1} & \textbf{51.5 ± 0.9} & \textbf{63.0 ± 6.1} & \textbf{51.3 ± 5.2} \\ \bottomrule
\end{tabular}
\caption{Main results: comparison of all methods on Earthquake and Wildfires datasets. 5 labeled data per class are used for all methods. Methods other than Supervised Baseline and TextMixUp leverages unlabeled data from the rest of training set. Results are averaged across three different runs. }
\label{tab3:main_result}
\end{table}

In Table \ref{tab3:main_result}, we summarize and compare the results of all methods in Earthquake and Wildfires datasets. We provide detailed analyses and discuss our findings.

\textbf{Leveraging unlabeled data achieves significantly better accuracy performance with limited labeled data. } 
As shown in Table \ref{tab3:main_result}, methods that leveraged unlabeled data, including PSL, PSL++ and CrisisMatch, significantly outperform the supervised baseline, by at least 5.4\% accuracy on the Wildfires dataset and 6.4\% accuracy on Earthquake dataset. In addition, CrisisMatch surpasses the supervised baselines by 9.5\% and 12.9\% accuracy on Earthquake and Wildfires datasets. This demonstrates that unlabeled data can be effectively used to boost performance when limited labeled data is available and thus can alleviate reliance on labeled data.

\textbf{Data augmentation and consistency regularization improve the performance of pseudo-labeling. }
We compare PSL with PSL++. PSL++ boosts the performance of PSL by 0.8\% accuracy and 4.6\% macro F1 on Wildfires dataset and 0.7\% accuracy and 2.8\% macro F1 on Earthquake dataset, justifying the effectiveness of data augmentation and consistency regularization on regularizing the model. 

\textbf{TextMixUp boosts the performance of the supervised baseline. }
Compared with the supervised baseline, TextMixUP increases 2.1\% accuracy, 2\% macro F1 on the Wildfire dataset, and 7\% accuracy, 5.1\% F1 on the Wildfires dataset. This attests that artificially generated mixed samples by TextMixUp alleviate overfitting and regularize the model to perform linearly between data.

\textbf{CrisisMatch achieves the best performance for fine-grained crisis tweet classification task. } 
The proposed CrisisMatch reaches the best performance of 63.4\% accuracy and 51.5\% F1 on Wildfires datasets, which obtains +9.5\% accuracy and +7.3\% F1 huge boosts than the supervised baseline. On the Earthquake dataset, CrisisMatch surpasses the supervised baseline by a large margin of 12.9\% accuracy and 9.9\% F1. Besides, CrisisMatch consistently performs better than TextMixMatch. These results demonstrate that our proposed CrisisMatch effectively incorporates different ideas and components to handle the fine-grained crisis tweet classification with limited unlabeled data.

%%%%%%%%%%%%%%%%

\subsection{4.5 Other Analysis}
\label{sec:other_analysis}

\textbf{Influence of number of labeled data}

\begin{figure}[H]
\centering
  \begin{minipage}[l]{\textwidth}
    \begin{minipage}[t]{0.5\textwidth}
      \centering
      \centerline{\includegraphics[width=\columnwidth]{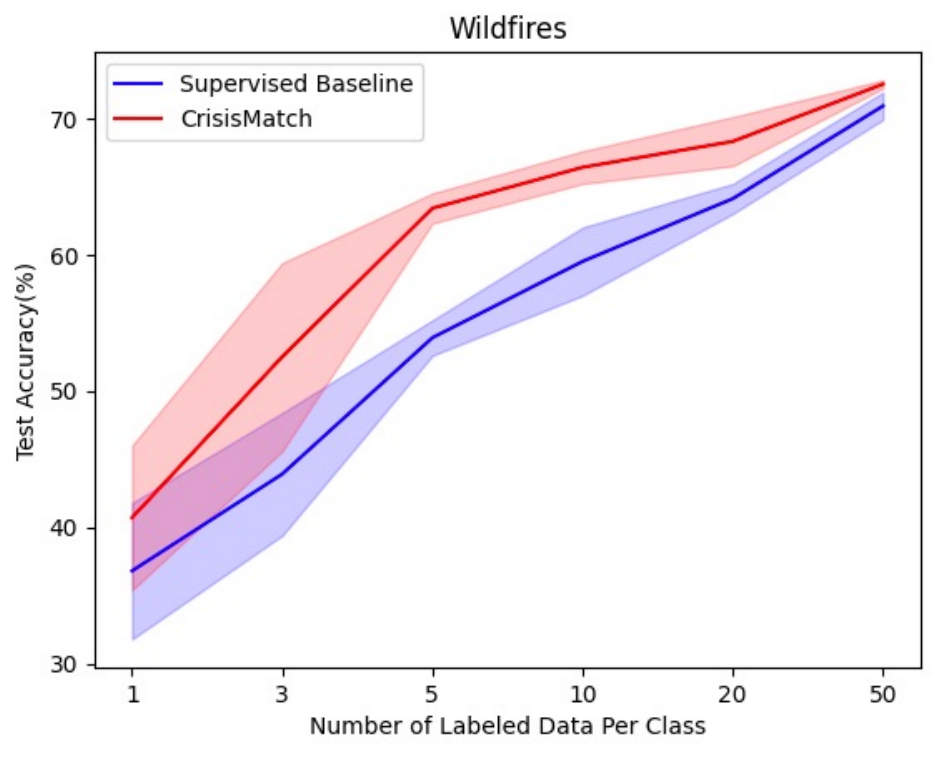}}
      % \captionof{figure}{...}
      % \label{fig:vary_cifar10}
    \end{minipage}\hfill
    \begin{minipage}[t]{0.5\textwidth}
      \centering
      \centerline{\includegraphics[width=\columnwidth]{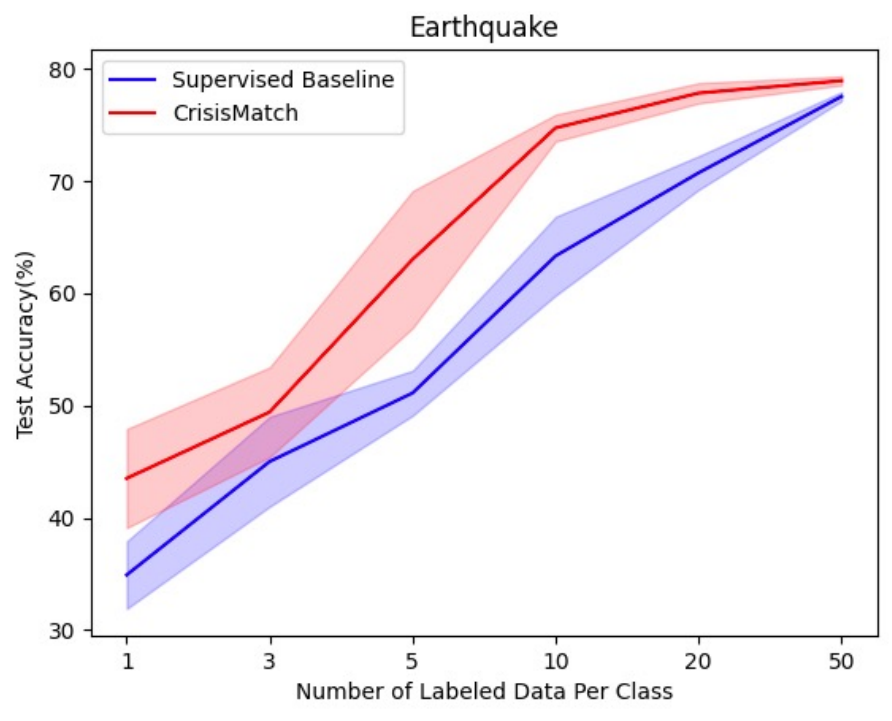}}
    %   \captionof{figure}{...}
    % \label{fig:vary_svhn}
    \end{minipage}
  \end{minipage}
\caption{Performance with varying number of labeled data per class.}
\label{fig1:influence_n_labeled}
\end{figure}

We evaluate our baseline and proposed method using accuracy and macro-F1 with a varying number of labeled data from 1 to 50. As shown in Table \ref{tab4:influence_n_labeled} and Figure \ref{fig1:influence_n_labeled}, CrisisMatch consistently obtains better accuracy than the supervised baseline in different settings, and also improves macro-F1 in most settings. As the number of labeled data increases, both the CrisisMatch and the supervised baseline achieve better performance in accuracy and macro-F1. For instance, CrisisMatch increases accuracy from 40.7\% to 72.5\% when the given number of labeled data per class is improved from 1 to 50. In general, the gap between CrisisMatch and the supervised baseline shrinks and both methods become more robust when more labeled data is provided, which we believe is because the supervision signal becomes sufficient.

% \usepackage{booktabs}
\begin{table}[H]
\centering
\begin{tabular}{@{}lllllll@{}}
\toprule
 & \multicolumn{6}{c}{Dataset: Wildfires} \\ \midrule
Accuracy & \multicolumn{1}{c}{1} & \multicolumn{1}{c}{3} & \multicolumn{1}{c}{5} & \multicolumn{1}{c}{10} & \multicolumn{1}{c}{20} & \multicolumn{1}{c}{50} \\ \midrule
Supervised Baseline & 36.8 ± 5.0 & 43.9 ± 4.5 & 53.9 ± 1.3 & 59.5 ± 2.5 & 64.1 ± 1.1 & 70.9 ± 1.0 \\
CrisisMatch & \textbf{40.7 ± 5.3} & \textbf{52.5 ± 6.9} & \textbf{63.4 ± 1.1} & \textbf{66.4 ± 1.2} & \textbf{68.3 ± 1.8} & \textbf{72.5 ± 0.3} \\ \midrule
Macro-F1 & \multicolumn{1}{c}{1} & \multicolumn{1}{c}{3} & \multicolumn{1}{c}{5} & \multicolumn{1}{c}{10} & \multicolumn{1}{c}{20} & \multicolumn{1}{c}{50} \\ \midrule
Supervised Baseline & 28.3 ± 2.9 & 35.5 ± 3.7 & 44.2 ± 1.8 & 50.2 ± 2.3 & 54.1 ± 1.3 & 61.4 ± 1.2 \\
CrisisMatch & \textbf{31.4 ± 3.1} & \textbf{39.3 ± 2.9} & \textbf{51.5 ± 0.9} & \textbf{52.9 ± 2.2} & \textbf{56.2 ± 1.5} & \textbf{62.1 ± 1.7} \\
\end{tabular}
\begin{tabular}{@{}lllllll@{}}
\midrule
 & \multicolumn{6}{c}{Dataset: Earthquake} \\ \midrule
Accuracy & \multicolumn{1}{c}{1} & \multicolumn{1}{c}{3} & \multicolumn{1}{c}{5} & \multicolumn{1}{c}{10} & \multicolumn{1}{c}{20} & \multicolumn{1}{c}{50} \\\midrule
Supervised Baseline & 34.9 ± 3.0 & 45.0 ± 4.0 & 51.1 ± 2.0 & 63.3 ± 3.5 & 70.7 ± 1.5 & 77.5 ± 0.4 \\
CrisisMatch & \textbf{43.5 ± 4.4} & \textbf{49.4 ± 4.0} & \textbf{63.0 ± 6.1} & \textbf{74.7 ± 1.2} & \textbf{77.8 ± 0.9} & \textbf{78.9 ± 0.4} \\ \midrule
Macro-F1 & \multicolumn{1}{c}{1} & \multicolumn{1}{c}{3} & \multicolumn{1}{c}{5} & \multicolumn{1}{c}{10} & \multicolumn{1}{c}{20} & \multicolumn{1}{c}{50} \\ \midrule
Supervised Baseline & 25.5 ± 3.7 & 36.5 ± 2.6 & 41.4 ± 1.9 & 51.9 ± 2.3 & 59.2 ± 1.5 & \textbf{65.7 ± 0.4} \\
CrisisMatch & \textbf{25.8 ± 5.0} & \textbf{38.4 ± 3.2} & \textbf{51.3 ± 5.2} & \textbf{60.8 ± 0.7} & \textbf{63.1 ± 0.7} & 65.5 ± 0.1 \\ \bottomrule
\end{tabular}
\caption{Influence of number of labeled data. Supervised baseline only uses labeled data while CrisisMatch utilizes unlabeled data from the rest of training set.}
\label{tab4:influence_n_labeled}
\end{table}

%%%%%%%%%%%%%%%%%%%%%%%%%
\textbf{Out-of-Domain Results}

% \usepackage{booktabs}
\begin{table}[H]
\centering
\begin{tabular}{@{}lllllll@{}}
\toprule
 & \multicolumn{6}{c}{Source: Earthquake, Target: Floods} \\ \midrule
Acc/Macro-F1 & \multicolumn{2}{c}{5} & \multicolumn{2}{c}{10} & \multicolumn{2}{c}{20} \\ \midrule
Supervised Baseline & 47.3 ± 7.1 & \multicolumn{1}{l|}{31.2 ± 5.1} & 58.9 ± 6.4 & \multicolumn{1}{l|}{39.8 ± 2.3} & 64.3 ± 3.6 & 45.9 ± 4.2 \\
CrisisMatch & \textbf{52.9 ± 13.1} & \multicolumn{1}{l|}{\textbf{32.6 ± 7.0}} & \textbf{66.9 ± 2.1} & \multicolumn{1}{l|}{\textbf{45.0 ± 0.5}} & \textbf{67.9 ± 0.8} & \textbf{47.8 ± 1.0} \\ \midrule
 & \multicolumn{6}{c}{Source: Wildfires, Target: Floods} \\ \midrule
Acc/Macro-F1 & \multicolumn{2}{c}{5} & \multicolumn{2}{c}{10} & \multicolumn{2}{c}{20} \\ \midrule
Supervised Baseline & \textbf{57.7 ± 3.4} & \multicolumn{1}{l|}{34.7 ± 1.5} & 61.0 ± 0.3 & \multicolumn{1}{l|}{40.7 ± 3.3} & 65.9 ± 0.6 & 46.4 ± 2.0 \\
CrisisMatch & 56.4 ± 2.8 & \multicolumn{1}{l|}{\textbf{37.6 ± 1.9}} & \textbf{66.5 ± 3.7} & \multicolumn{1}{l|}{\textbf{44.7 ± 3.3}} & \textbf{70.6 ± 0.1} & \textbf{49.3 ± 2.9} \\ \bottomrule
\end{tabular}
\caption{Out-of-domain results.}
\label{tab5:out-of-domain}
\end{table}

We investigate the performance of our model on data from out-of-domain (i.e., the distribution of testing data is different from the distribution of training data). Specifically, we use the Earthquake or Wildfires as training data, with each class having 5, 10, or 20 labeled data, and test the model on the Floods dataset. As shown in Table \ref{tab5:out-of-domain}, our proposed CrisisMatch generally increases both accuracy and macro-F1 over the supervised baseline. For instance, given 10 labeled data per class on the Earthquake dataset, CrisisMatch outperforms the supervised baseline by 8\% accuracy and 5.2\% macro-F1 score. This observation confirms the robustness of our proposed CrisisMatch on out-of-domain data.

\textbf{Entropy minimization: sharpening vs. hard pseudo-labeling. }

We experimented with two different approaches to minimize entropy. CrisisMatch with hard pseudo-labeling empirically performs slightly better than sharpening with soft labels. For instance, compared to CrisisMatch with hard pseudo-labeling, CrisisMatch with sharpening decreases the performance by 2.8\% macro-F1 on Wildfires and 3.5\% accuracy on Earthquake dataset. These results show that hard pseudo-labeling is a more effective method in entropy minimization for our fine-grained crisis tweet classification task.

% \usepackage{booktabs}
\begin{table}[H]
\centering
\begin{tabular}{@{}lllll@{}}
\toprule
Datasets & \multicolumn{2}{c}{Wildfires} & \multicolumn{2}{c}{Earthquake} \\ \midrule
Methods & Accuracy & Macro-F1 & Accuracy & Macro-F1 \\ \midrule
Sharpening & 62.0 ± 7.5 & 48.5 ± 5.4 & 59.9 ± 3.1 & 47.5 ± 3.2 \\
Hard pseudo-labeling & \textbf{63.0 ± 6.1} & \textbf{51.3 ± 5.2} & \textbf{63.4 ± 1.1} & \textbf{51.5 ± 0.9} \\ \bottomrule
\end{tabular}
\caption{Entropy minimization: sharpening vs. hard pseudo-labeling.}
\label{tab6:entropy_min}
\end{table}

\subsection{4.6 Ablation Study}
Since CrisisMatch comprises various existing mechanisms, we conduct extensive ablation studies to show the effectiveness of each component. 

% \textbf{Remove different parts from CrisisMatch}

% \usepackage{booktabs}
\begin{table}[H]
\centering
\begin{tabular}{lll}
\toprule
\textbf{Methods} & \textbf{Accuracy} & \textbf{Macro-F1} \\ \midrule
CrisisMatch & 63.4 ± 1.1 & 51.5 ± 0.9 \\ \midrule
- Unlabeled Data & 57.6 ± 3.6 & 47.4 ± 3.2 \\
- Consistency Regularization & 62.7 ± 3.0 & 49.3 ± 4.2 \\
- TextMixUp & 59.3 ± 2.1 & 48.6 ± 1.4 \\
- All (Supervised Baseline) & 53.9 ± 1.3 & 44.2 ± 1.8 \\ \bottomrule
\end{tabular}
\caption{Ablation study on Wildfires dataset with 5 labeled data per class.}
\label{tab7:ablation_study}
\end{table}

As illustrated in Table \ref{tab7:ablation_study}, we empirically study the performance of CrisisMatch after removing each of its components at a time. One may notice that the performance drops after stripping each part, demonstrating that all components contribute to CrisisMatch to achieve better results. Furthermore, removing unlabeled data reduces the performance most, suggesting that CrisisMatch can effectively leverage unlabeled data to train a better model. 
Removing TextMixUp results in the second-largest decrease in performance. This implies the effectiveness of TextMixUp in providing data augmentation and regularizing the model. In addition, the decrease resulting from removing consistency regularization illustrates the importance of encouraging the model to perform consistently for different augmentations of data.

% \section{5 Limitations}

\section{5 Conclusion and Future Work}
In this paper, we studied several methods on how to effectively leverage unlabeled data for disaster tweet classification in the semi-supervised few-shot setting, where there are only a few labeled data per class but large amounts of unlabeled data are available. Concretely, we introduced different variants of pseudo-labeling by introducing data augmentation, entropy minimization and consistency regularization. Besides, we studied TextMixUp and proposed our algorithm CrisisMatch, which further integrated TextMixUp into pseudo-labeling. Experimental results show that CrisisMatch can surpass the supervised baseline by a significant margin and demonstrate its effectiveness in utilizing unlabeled data. In the future, we plan to deal with the existing data-imbalanced issue and explore adaptive thresholds for different classes on a wider range of datasets and tasks.

\printbibliography[heading=bibliography]
\end{document}